# Applying Faster R-CNN for Object Detection on Malaria Images

Jane Hung
Massachusetts Institute of Technology
jyhung@mit.edu

Deepali Ravel*
Harvard T.H. Chan School of Public Health (HSPH)

Stefanie C.P. Lopes*
Instituto Leônidas e Maria Deane, Fundação Oswaldo Cruz (FIOCRUZ); Fundação de Medicina Tropical Dr. Heitor Vieira Dourado, Gerência de Malária

Gabriel Rangel*
Harvard T.H. Chan School of Public Health (HSPH)

Odailton Amaral Nery
Universidade de São Paulo

Benoit Malleret
Department of Microbiology & Immunology, Yong Loo Lin School of Medicine, National University of Singapore; Singapore Immunology Network (SIgN), Agency for Science & Technology

Francois Nosten
Shoklo Malaria Research Unit, Mahidol-Oxford Tropical Medicine Research Unit, Faculty of Tropical Medicine, Mahidol University; Centre for Tropical Medicine and Global Health, Nuffield Department of Medicine, University of Oxford

Marcus V.G. Lacerda
Fundação de Medicina Tropical Dr. Heitor Vieira Dourado, Gerência de Malária

Marcelo U. Ferreira
Institute of Biomedical Sciences, University of São Paulo

Laurent Rénia
Singapore Immunology Network (SIgN), Agency for Science & Technology (A*STAR)

Manoj T. Duraisingh
Harvard T.H. Chan School of Public Health (HSPH)

Fabio T.M. Costa
Department of Genetics, Evolution, Microbiology and Immunology, University of Campinas

Matthias Marti
Wellcome Trust Center for Molecular Parasitology, University of Glasgow

Anne E. Carpenter
Imaging Platform, Broad Institute of Harvard and MIT
anne@broadinstitute.org
https://www.broadinstitute.org/~anne/

**Abstract**
*Deep learning based models have had great success in object detection, but the state of the art models have not yet been widely applied to biological image data. We apply for the first time an object detection model previously used on natural images to identify cells and recognize their stages in brightfield microscopy images of malaria-infected blood. Many micro-organisms like malaria parasites are still studied by expert manual inspection and hand counting. This type of object detection task is challenging due to factors like variations in cell shape, density, and color, and uncertainty of some cell classes. In addition, annotated data useful for training is scarce, and the class distribution is inherently highly imbalanced due to the dominance of uninfected red blood cells. We use Faster Region-based Convolutional Neural Network (Faster R-CNN), one of the top performing object detection models in recent years, pre-trained on ImageNet but fine tuned with our data, and compare it to a baseline, which is based on a traditional approach consisting of cell segmentation, extraction of several single-cell features, and classification using random forests. To conduct our initial study, we collect and label a dataset of 1300 fields of view consisting of around 100,000 individual cells. We demonstrate that Faster R-CNN outperforms our baseline and put the results in context of human performance.*

## 1. Introduction

Biology contains a multitude of problems made for object detection. Although there has been a lot of interest in deep learning based models and their success in object detection, the state of the art models from competitions like ImageNet Large Scale Visual Recognition Challenge (ILSVRC)[1] and MS-COCO[2] have not yet been widely applied to biological image data. We are interested in using object detection to identify cells and recognize their categories for diseases such as malaria, where manual inspection of microscopic views by trained experts remains the gold standard. A robust solution would allow for automated single cell classification and counting and would provide enormous benefits due to faster and more accurate quantitative results without human variability[3].

Object detection of cells in brightfield microscopy images presents special challenges. Like natural images, microscopy images of malaria-infected blood have variations in illumination from the microscope, in cell shape, density, and color from variations in sample preparation, and have objects of uncertain class (even for experts). However, unlike natural images, there is a dearth of annotated data useful for training because of the scarcity of experts, and the class distribution is inherently highly imbalanced due to the dominance of uninfected red blood cells (RBCs).

Previous attempts to automate the process of identifying and quantifying malaria[4–6] have used complex workflows for image processing and machine learning classification using features from a predetermined set of measurements (intensity, shape, and texture). However, none of these methods have gained major traction because of a lack of generalizability and difficulty of replication, comparison, and extension. Algorithms cannot be reimplemented with certainty nor extended because authors do not generally make functioning code available. Authors also rarely make their image sets available, which precludes replication of results. The lack of a standard set of images nor standard set of metrics used to report results has impeded the field.

For our task of detecting individual cells and their classes, we choose to use a deep learning based framework called Faster Region-based Convolutional Neural Network (Faster R-CNN)[7] because R-CNN[8] and its successors[7,9] have been the basis of the top performing object detection models in recent years. In contrast to previous methods, this one avoids the task of segmentation and does not rely on general features for classification. As a baseline, we develop a traditional approach consisting of cell segmentation and extraction of several single-cell level features, followed by classification using random forests. To conduct our initial study, we collect a novel dataset: 1300 microscopy images consisting of 100,000 individual cells.

## 2. Data

Our data came from three different labs' ex vivo samples of *P. vivax* infected patients in Manaus, Brazil, and Thailand. The Manaus and Thailand data were used for training and validation while the Brazil data were left out as our test set. Blood smears were stained with Giemsa reagent, which attaches to DNA and allow experts to inspect infected cells and determine their stage.

All non-RBC objects were annotated (boxed and labeled) by an expert malaria researcher (Figure 1). Seven labels used to cover possible cell types of interest: RBC, leukocyte, gametocyte, ring, trophozoite, and schizont. RBCs and leukocytes are uninfected cell types normally found in blood. Infected cells can develop either sexually (as gametocytes) or asexually (rings, trophozoites, then schizonts). Some cells were marked as difficult when not clearly in one of the classes, but those marked difficult were ignored in training. The data is also naturally imbalanced among the object classes. RBCs clearly dominate with about 97% of 100,000 total cells.

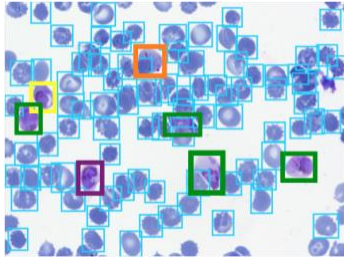

Figure 1. Example image with annotations, with colors representing different class labels.

## 3. Establish a Baseline with Traditional Method

We established a baseline performance level of an automated detection and classification model using the traditional approach of segmentation followed by machine learning. By feeding in full sized training set images into open source image processing software CellProfiler[14], which is geared towards biological images, we segmented cells and obtained about 300 intensity, shape, and texture feature measurements for each cell. Since the segmentations may not completely match the ground truth (Figure 2), for each segmented object we found the ground truth object with the most overlap and call it a match if the intersectional area over area of union (IoU) exceeded 0.4. Otherwise, the objects were mis-segmentations and not used for training. We took measurements of each cell and corresponding cell labels to train a machine learning classifier, Random Forest with n=1000 trees implemented with scikit-learn[15]. Balancing was done by adjusting class weights.

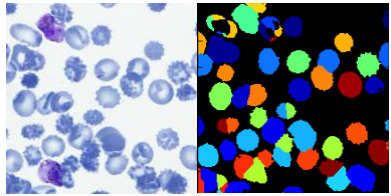

Figure 2. Example segmentation (right is after segmentation). Some mis-segmentations are due to one cell being split into multiple and some are due to multiple cells being seen as one.

## 4. Deep Learning

Rather than use the full sized images for training the deep learning models, we took 448x448 crops of each full image to augment the number of training examples and cut down on training time. Enough crops of each full image were taken such that the number of cells contained in the crops was at least twice the number of cells contained in the full sized image up to a maximum of 100 crops.

As is, the training set is highly imbalanced towards RBCs, so to create a more balanced training set, we rotated crops containing underrepresented classes by 90 degrees, which augmented underrepresented cell counts by roughly 4 times and removed crops containing only RBCs.

### 4.1. Faster R-CNN

Figure 3. Faster R-CNN is a network for object detection that takes an image and outputs bounding boxes around objects of interest with class labels. It contains a Region Proposal Network that acts as an attention network that identifies the regions the classifier should consider[11].

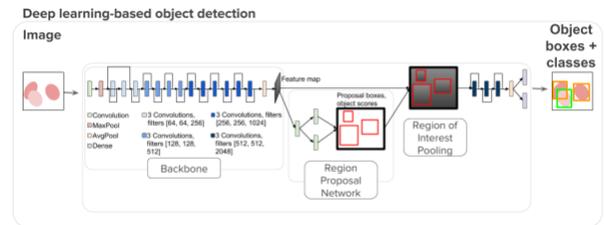

Faster R-CNN[11] became the state of the art in object detection when it was released in 2015. It takes an image as input and runs it through 2 modules: the first uses a region proposal network (RPN) that proposes object regions and the second is a Fast R-CNN object detector that classifies the region proposals[13] (Figure 3). To save time, RPN and Fast R-CNN share convolutional layers.

## 4.2. Two Stage Detection and Classification

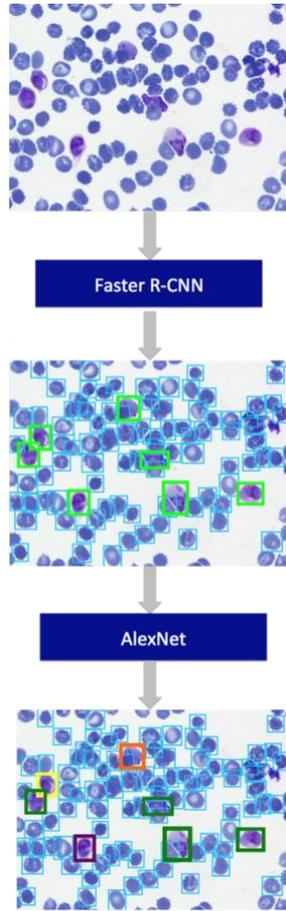

Figure 4. Overview of how our two stage deep learning model for detection and classification is applied to images (i.e. test phase). A full sized image is fed into Faster R-CNN to detect objects and label them as RBC or other. The objects labeled as other are sent to AlexNet or another CNN to undergo more fine-grained classification.

Our model detects and classifies in two stages (Figure 4). We choose a two-stage approach because the subtle differences between infected classes are difficult to distinguish when the large difference between RBCs and other cells is present. In stage one, object detection framework Faster R-CNN identifies bounding boxes around objects and classifies them as RBC or other (including infected cells and leukocytes). Faster R-CNN uses a convolutional neural network- here we use the AlexNet architecture with 7 layers- to jointly detect objects (vs. non-objects) and assign classes to the objects. In stage two, the detections from stage one labeled as other (non-RBC) are fed into AlexNet to obtain a 4096 dimensional feature vector used to classify them into more fine grained categories.

## 4.3. Implementation

All deep learning models have been pre-trained with the natural image dataset ImageNet[16] and fine-tuned with our training data. The images in the training set were randomly split into a training and validation set for tuning the learning rate and determine early stopping.

In the two-stage model, additional augmentation was done to individual cell images. This included rotations, flips, horizontal and vertical shifts, color channel shifts, and scale shifts.

## 5. Results and Discussion

### 5.1. Baseline

Using a traditional segmentation plus machine learning method as a baseline, we see that the model attains 50% accuracy (disregarding background, RBCs, and difficult cells) compared to the ground truth matched segmentation objects (Figure 5). Note that this is an overestimate of the accuracy compared to the true ground truth as it does not include mis-segmentation error.

|  | Model Count | Ground Truth Matched Count | Ground Truth Count |
| --- | --- | --- | --- |
| RBC | 19561 | 19181 | 19604 |
| trophozoite | 521 | 538 | 561 |
| schizont | 6 | 26 | 28 |
| ring | 4 | 81 | 88 |
| gametocyte | 20 | 76 | 75 |
| leukocyte | 23 | 30 | 28 |
| difficult | 0 | 217 | 218 |

Figure 5: Table of predicted counts, counts from matching segmented objects to ground truth, and ground truth counts.

### 5.2. Deep Learning: One-Stage Classification

The results of using Faster R-CNN to detect and classify all objects are shown in Figure 6. The accuracy is 59% (disregarding background, RBCs, and difficult cells). From counts alone, it is difficult to see whether the model is finding distinguishing features between infected classes. To easily visualize the feature vector that Faster R-CNN uses to do classification, we display a 2D t-SNE plot[18] in Figure 6b. t-SNE plots can be used to visualize high dimensional data in 2D in a way that maintains local structures. Pairs of points are given joint probabilities based on their distance and the Kullback-Leibler divergence between the probabilities is minimized.

There is a clear separation between RBCs (in blue) and the other classes, but within the cluster of infected cells, there is a lack of separation between the different stages. This shows that the features learned through training can clearly distinguish most of the RBCs from the other types, but they are not sufficient to distinguish the subtle differences between infected stages. The accuracy has been greatly affected by cells being identified as multiple cell types, which further indicates confusion of the model. The model might not be able to classify the cells into more fine-grained categories because of the extreme class imbalance. It is not clear how to do further class balancing within this framework, but once individual cells are identified, we can utilize techniques used in image classification.

### 5.3. Deep Learning: Two-Stage Classification

Our two-stage model uses Faster R-CNN to detect all objects and classify them as RBC or not and a separate image classifier to make detailed classifications of the detections that Faster R-CNN labeled as not-RBC.

The results are shown in Figure 7. The total accuracy is 98% (disregarding background, RBCs, and difficult cells), which is a significant improvement over the one stage method. The t-SNE plot in Figure 7b shows clear clusters that differentiate the classes, even those marked as difficult. The model can use what it has learned from training to confidently classify cells that humans are confused by and normally ignore. This large increase in accuracy is likely due to changes in the training process: learning rate adjustments based on subsets of training data, increase of the number of layers, and additional augmentation.

a)

| | Model Count | Ground Truth Count |
|---|---|---|
| RBC | 19112 | 19604 |
| trophozoite | 664 | 561 |
| schizont | 39 | 28 |
| ring | 227 | 88 |
| gametocyte | 74 | 75 |
| leukocyte | 49 | 28 |
| difficult | 0 | 218 |

b)

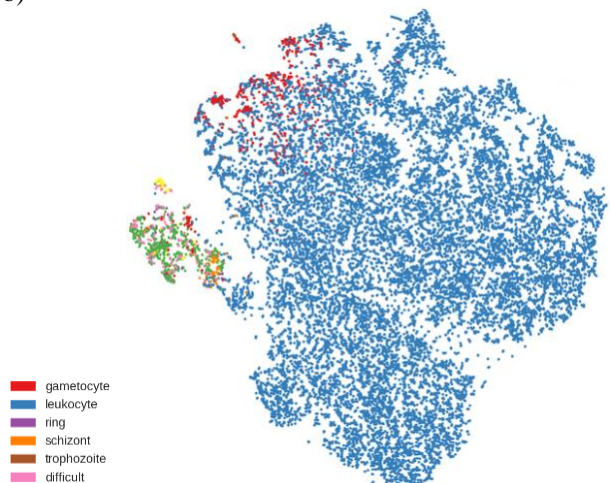

Figure 6. a) Table of predicted counts with threshold = 0.65 and ground truth counts. b) t-SNE plot.

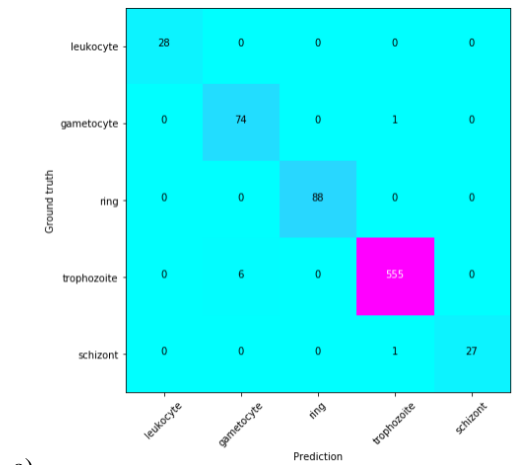

a)

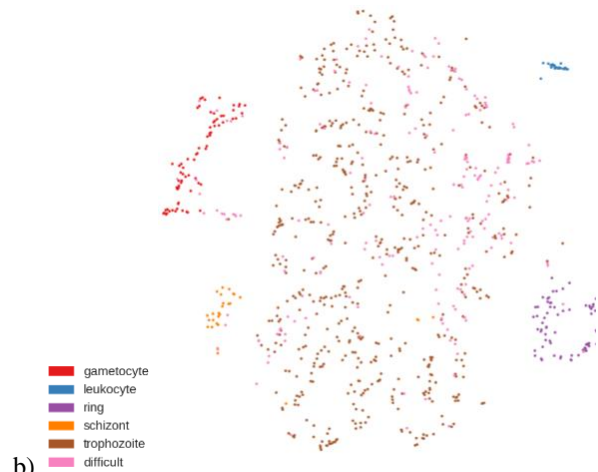

b)

Figure 7. a) Confusion matrix of predicted vs. ground truth in the classification stage. b) t-SNE plot after 2-stage classification.

## 5.4. Comparison with Humans

In order to put the results in context, we compare the results of our deep learning model to human annotators. Two expert annotators from two different *P. vivax* labs annotated all the images from the test set independently, and the results are shown in Figure 8. The total accuracy of the non-difficult infected cells is 72%, which is less than the accuracy of our two-stage classification model. This shows the model's ability to identify cells as well as an expert human for cases where the humans are sure about the classification. In ambiguous cases, an automated system would be even more useful for humans.

|  | Annotator 1 Count | Annotator 2 Count | F1 score (%) |
|---|---|---|---|
| trophozoite | 561 | 437 | 82 |
| schizont | 28 | 98 | 44 |
| ring | 88 | 40 | 67 |
| gametocyte | 75 | 242 | 63 |
| leukocyte | 28 | 23 | 92 |
| difficult | 218 | 119 | -- |

Figure 8. Table of counts for two expert annotators and the model predictions.

## 5.5. Future Work

The end goal of this project is to develop a framework that can help researchers automatically classify and stage cells from a field of view image and identify features differentiating the infected stages. Further validation of our model needs to be done. We intend to test the model on more reliable ground truth (like samples with parasites with more synchronized growth) and test for robustness by testing on samples prepared in a different lab. We also intend to create an online tool where images can be run through the model, relevant results can be displayed, and annotated data from the community can be collected and incorporated into future iterations of the model.

# References

1. Russakovsky, O. *et al.* ImageNet Large Scale Visual Recognition Challenge. *Int. J. Comput. Vis.* **115,** 211–252 (2015).
2. Lin, T.-Y. *et al.* Microsoft COCO: Common Objects in Context. in *Computer Vision – ECCV 2014* (eds. Fleet, D., Pajdla, T., Schiele, B. & Tuytelaars, T.) 740–755 (Springer International Publishing, 2014).
3. Mavandadi, S. *et al.* A mathematical framework for combining decisions of multiple experts toward accurate and remote diagnosis of malaria using tele-microscopy. *PLoS One* **7,** e46192 (2012).
4. Linder, N. *et al.* A malaria diagnostic tool based on computer vision screening and visualization of Plasmodium falciparum candidate areas in digitized blood smears. *PLoS One* **9,** e104855 (2014).
5. Díaz, G., González, F. A. & Romero, E. A semi-automatic method for quantification and classification of erythrocytes infected with malaria parasites in microscopic images. *J. Biomed. Inform.* **42,** 296–307 (2009).
6. Tek, F. B., Dempster, A. G. & Kale, İ. Parasite detection and identification for automated thin blood film malaria diagnosis. *Comput. Vis. Image Underst.* **114,** 21–32 (2010).
7. Ren, S., He, K., Girshick, R. & Sun, J. Faster R-CNN: Towards Real-Time Object Detection with Region Proposal Networks. *arXiv [cs.CV]* (2015).
8. Girshick, R., Donahue, J., Darrell, T. & Malik, J. Region-Based Convolutional Networks for Accurate Object Detection and Segmentation. *IEEE Trans. Pattern Anal. Mach. Intell.* **38,** 142–158 (2016).
9. Girshick, R. Fast r-cnn. in *Proceedings of the IEEE International Conference on Computer Vision* 1440–1448 (2015).
10. Kamentsky, L. *et al.* Improved structure, function and compatibility for CellProfiler: modular high-throughput image analysis software. *Bioinformatics* **27,** 1179–1180 (2011).
15. Pedregosa, F. *et al.* Scikit-learn: Machine Learning in Python. *J. Mach. Learn. Res.* **12,** 2825–2830 (2011).
16. Deng, J. *et al.* ImageNet: A large-scale hierarchical image database. in *Computer Vision and Pattern Recognition, 2009. CVPR 2009. IEEE Conference on* 248–255 (2009).
17. Dwork, C. *et al.* STATISTICS. The reusable holdout: Preserving validity in adaptive data analysis. *Science* **349,** 636–638 (2015).
18. Maaten, L. van der & Hinton, G. Visualizing Data using t-SNE. *J. Mach. Learn. Res.* **9,** 2579–2605 (2008).
19. Russell, B. *et al.* A reliable ex vivo invasion assay of human reticulocytes by *Plasmodium vivax*. *Blood* **118**, 74-81 (2011).